\documentclass[a4paper,conference]{IEEEtran}
\usepackage{cite}
\usepackage[pdftex]{graphicx}
\usepackage[caption=false,font=normalsize,labelfont=sf,textfont=sf]{subfig}
\usepackage[hyphens]{url}

\usepackage{helvet}
\usepackage{courier}

\usepackage{multirow}
\usepackage{diagbox}

\begin{document}
\title{Multi-modal Identification of State-Sponsored Propaganda on Social Media}

\author{\IEEEauthorblockN{Xiaobo Guo}
\IEEEauthorblockA{Department of Computer Science, 
Dartmouth College\\
Hanover, New Hampshire 03755\\
Email: xiaobo.guo.gr@dartmouth.edu}
\and
\IEEEauthorblockN{Soroush Vosoughi}
\IEEEauthorblockA{Department of Computer Science,
Dartmouth College\\
Hanover, New Hampshire 03755\\
Email: soroush.vosoughi@dartmouth.edu}}

\maketitle

\begin{abstract}
  The prevalence of state-sponsored propaganda on the Internet has become a cause for concern in the recent years. While much effort has been made to identify state-sponsored Internet propaganda, the problem remains far from being solved because the ambiguous definition of propaganda leads to unreliable data labelling, and the huge amount of potential predictive features causes the models to be inexplicable. This paper is the first attempt to build a balanced dataset for this task. The dataset is comprised of propaganda by three different organizations across two time periods. A multi-model framework for detecting propaganda messages solely based on the visual and textual content is proposed which achieves a promising performance on detecting propaganda by the three organizations both for the same time period (training and testing on data from the same time period) (F1=0.869) and for different time periods (training on past, testing on future) (F1=0.697). To reduce the influence of false positive predictions, we change the threshold to test the relationship between the false positive and true positive rates and provide explanations for the predictions made by our models with visualization tools to enhance the interpretability of our framework. Our new dataset and general framework provide a strong benchmark for the task of identifying state-sponsored Internet propaganda and point out a potential path for future work on this task.
\end{abstract}

\section{Introduction}
    Propaganda, a form of communication that attempts to achieve the response that furthers the desired intent of the propagandist\cite{jowett2018propaganda_1}, is difficult to distinguish from persuasion, as they have many attributes in common. However, three main factors help distinguish between them: the purpose, the parties spreading the message, and the veracity (i.e. whether the news is fake or based on fact). While traditional propaganda spread through radio, television, and other media platforms is relatively easy to recognize by tracking the dissemination path, propaganda spread via Internet is hard to identify as the path is not always clear. One of the most harmful types of Internet propaganda is state-sponsored propaganda. This is because these actors have large resources at their disposal that they can use to sway public opinion by flooding social media with their messages. The situation is made more complex by the practice of astroturfing, which refers to the practice of cajoling readers into believing that the content comes from grassroots individuals instead of the actual behind-the-scenes sponsors.
    
    To detect state-sponsored Internet propaganda and minimize its harmful effects, much research in the past several years has been conducted to identify commonly used technologies (e.g. bots, digital marketing) and strategies (e.g. misinformation, false news). Early approaches were based on various handcrafted features generated from the content of social media posts and the social context of users to identify misinformation \cite{jin2014news}, detect bot behavior \cite{kudugunta2018deep,ferreira2019uncovering}, and verify facts \cite{jin2016news,jin2016novel}. More recent approaches, inspired by the great success of neural networks, include various deep learning methods and have achieved promising results \cite{ma2018rumor,yu2017convolutional,nguyen2019fake,chen2018call}. However, these technologies and strategies (misinformation, bots, etc) are not exclusive to propaganda. Therefore, we aim to solve the problem of identifying state-sponsored Internet propaganda directly instead of focusing on these technologies and strategies.
    
    To identify state-sponsored propaganda on social media, four challenges need to be solved. The first is to collect and label data. Generating high-quality labels for such a dataset is a difficult task as even human annotators cannot easily distinguish propaganda from ``geninue'' messages. The second is to propose a framework able to identify propaganda from different organizations and on different topics at an early stage. The third is to reduce the potential damage of type \uppercase\expandafter{\romannumeral2} error (i..e, false positives). The last is to explore the generalizability of the features and the models to aid in further proposing models with more generalizability.

\begin{figure}[!htb]
        \centering
            \subfloat[]{
            \includegraphics[width=0.3\columnwidth]{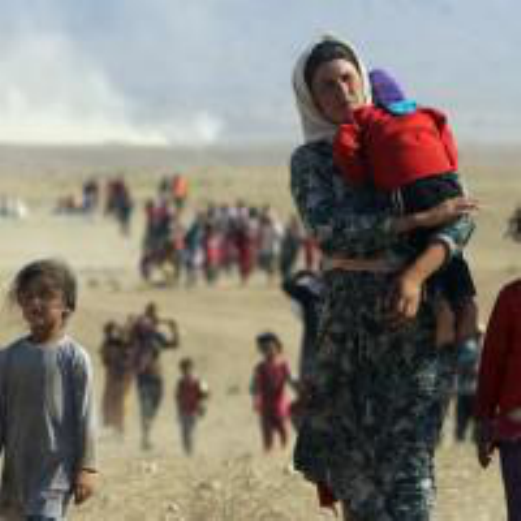}
            }
            \subfloat[]{
            \includegraphics[width=0.3\columnwidth]{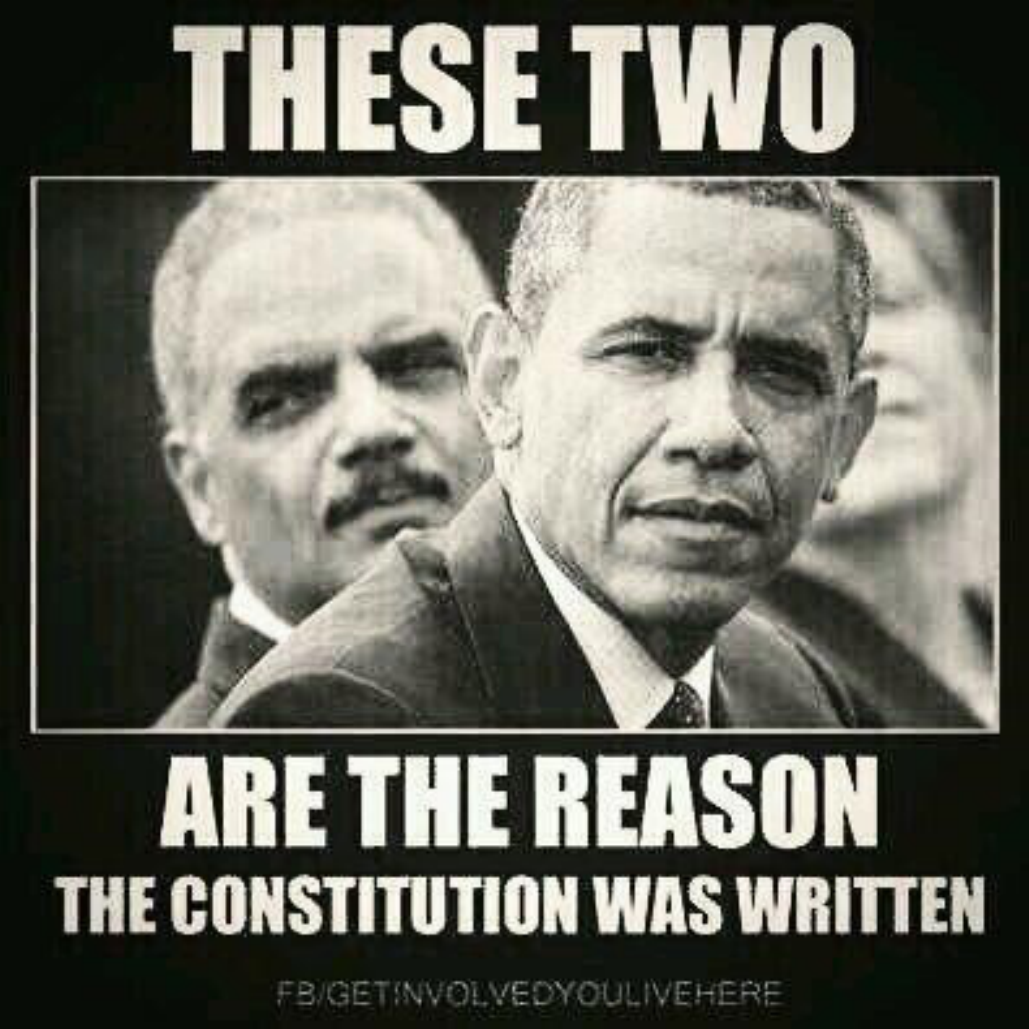}
            }
            \subfloat[]{
            \includegraphics[width=0.3\columnwidth]{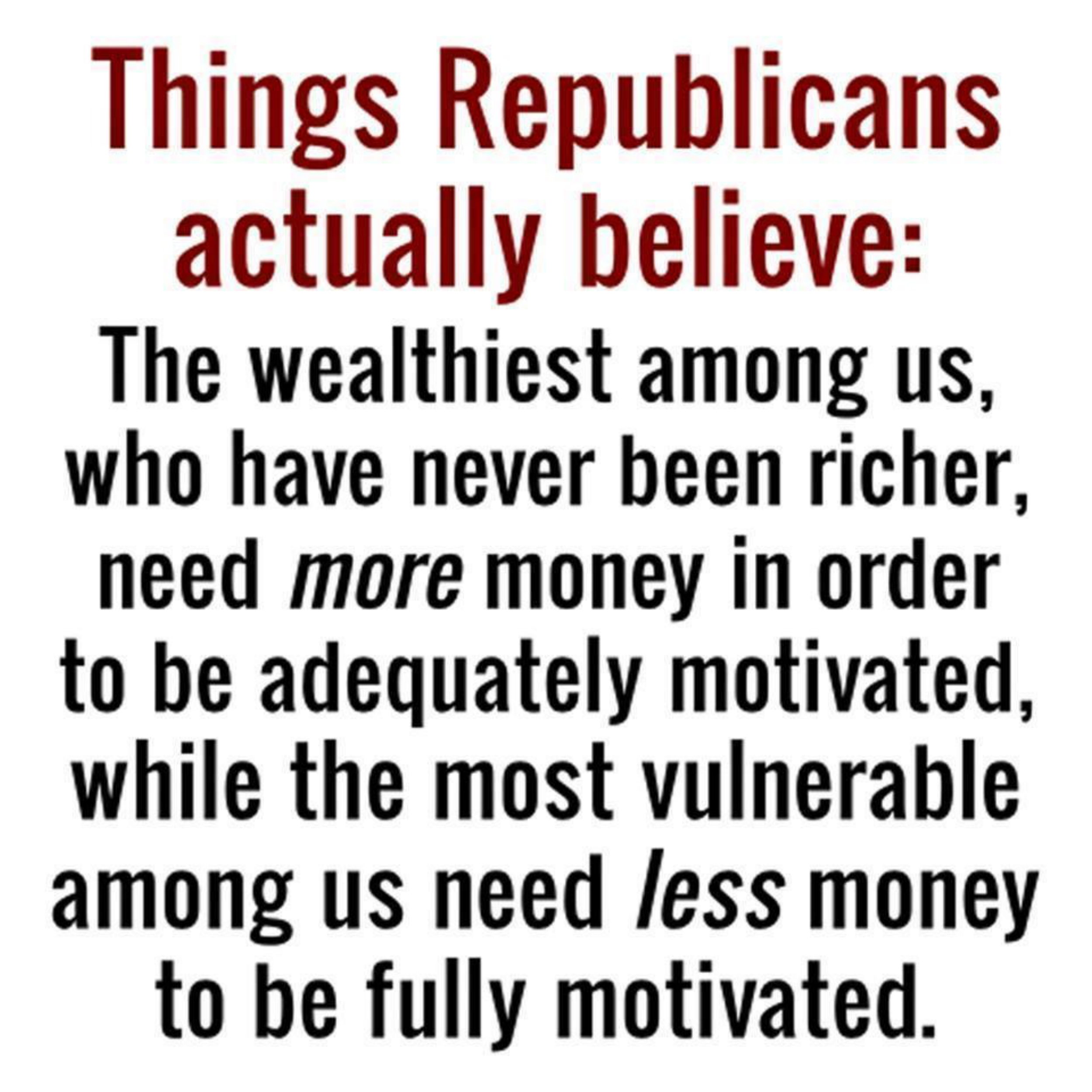}
            }
        \caption{Examples of images on Twitter. (a) is pure image, (b) is the combination of image and text and (c) is pure text.}
        \label{fig:image_example}
    \end{figure}

    In this paper, to solve the challenge of data labelling, we use a set of tweets labelled by Twitter\footnote{\url{https://blog.twitter.com/en_us/topics/company/2019/new-disclosures-to-our-archive-of-state-backed-information-operations.html}} as belonging to accounts reliably linked to state-sponsored trolls. These data serve as ground truth for state-sponsored propaganda. We combine this dataset with a dataset of random background tweets from the same time period and around the same topics that serve as our negative samples. To address the challenge of providing a general framework that can identify state-sponsored propaganda on social media, we make use of textual and visual content. Since the visual content of tweets includes primarily images and few videos, our framework makes use of only images. As shown in Figure \ref{fig:image_example}, the visual content of tweets is different from traditional images because it may be the combination of text and objects or even text only, which prompts us to test models focusing on different aspects of visual content. The textual content of tweets is a combination of words, hashtags, and the structural information (e.g. mentions and URLs). To further understand the textual content, they are processed in different ways to mask potential predictive features (such as removing hashtags). To tackle the challenge of reducing the potential damage of type \uppercase\expandafter{\romannumeral2} error, we explore the relationship between false positive and true positive rates and use visualization tools to highlight the regions of the visual and textual content that are used by our model for labelling. To study the generalizability of our features, we train and test models on datasets from a combination of different organizations to learn which features are generalizable.

    In particular, our work makes the following four contributions:
    \begin{itemize}
        \item We build the first balanced dataset for the work of identifying state-sponsored propaganda on Twitter which covers three different organizations during two time periods.
        \item To the best of our knowledge, we are the first to present a general multi-model framework for identifying state-sponsored propaganda on social media based solely on the content of messages and nothing else. 
        \item To reduce the potential damage of false positive predictions and add transparency, we use visualization tools to produce ``visual explanations'' for the predictions and conduct error analysis.
        \item We discover features extracted from the content can be divided into three types based on their generalizability: fundamental features, organizational features, and topic feature. By inhibiting part of these features, the rest will take more weights in models.
    \end{itemize}
    
\section{Related Work}
    Conventional works leverage handcrafted features from user credibility and content at post level \cite{barron2019proppy,ferreira2019uncovering}, at event level \cite{zhao2015enquiring,jin2016novel}, or aggregating from post level to event level \cite{jin2014news}. Some other works also adopt more complicated but efficient handcrafted features, such as conflict view points \cite{jin2016news}, temporal properties \cite{kwon2013prominent}, user feedback \cite{tschiatschek2018fake}, and dissemination patterns \cite{wu2015false}. However, handcrafted features may not be able to cover all potential features, and the high-level interactions among different features are likely to be missed.
    
    Others have looked at the application of different model architectures to neighbouring domains (such as rumor detection, fake news detection, etc). Nguyen et al. \cite{nguyen2019fake} used Deep Markov random fields for fake news detection, Ma et al. \cite{ma2018detect} used multi-task networks for rumor detection, Zhang et al. \cite{zhang2019multi}, and Wang et al. \cite{wang2018eann} used multi-model networks for rumor and fake news detection respectively.
    
    Our feature and model design is informed by the mentioned prior work. Specifically, we leverage features that are hard to manipulate and use multi-model neural network to capture relationship among them to identify state-sponsored Internet propaganda.
    
\section{Data}
    We build a balanced dataset that includes tweets of three different state-sponsored propagandists covering two time periods. This dataset labels state-sponsored Internet propaganda at the post level and can be used to test both whether the framework is feasible across different organizations and whether the model trained on one time period of one organization works for other organizations or time periods. The first time period is from Apr 1, 2015 to Feb 28, 2016 and is split into training, validation and ``continuous" test data by time (by time means that test data is comprised of posts made after the training posts). The second time period is from October 1, 2016 to June 30, 2017 and is used as ``delay'' test data. We choose these two time periods because 
   they are the most active time periods for the propagandists we choose in terms of the number of tweets they publish and retweet.
    
    Continuous and delay here refer to whether the test data is in the same temporal vicinity of the the training data or much further in the future. Both are important test cases for studying the generalizability of our model.
    
    To build our dataset of state-sponsored Internet propaganda, we use tweets that are believed to be from potentially state-sponsored information operations on Twitter. Starting from October 2018, Twitter has started to routinely disclose datasets of information operations which can be reliably linked to state-sponsored trolls (using Twitter's own internal analytics). We focus on three organizations: the data released in October 2018 from the \emph{Internet Research Agency (IRA)}, in October 2018 from \emph{Iranian-backed (Iranian)} accounts, and in January 2019 from \emph{Russian-linked (Russian)} accounts (note that even though IRA is also a Russian agency, Twitter specifically separates these datasets as they are presumably operated by different organizations within Russia) . All tweets from state-sponsored trolls are regarded as state-sponsored regardless of whether they are original tweets or retweets. This is because we assume that both tweets originally published by state-sponsored trolls and those retweeted by the trolls all push the same agenda on important topics. 
    
    Only propaganda tweets written in English are kept and are standardized by replacing URLs with 'URL', and removing other non-content characters ('rt','RT','rt @username', and 'RT @username'). Since we are interested in detecting propaganda based on content, we remove tweets with duplicate content. We do this at the text and image level. At the text level, we compare tweets using the standardized textual content. At the image level, we generate a hash\footnote{image hash is generated using \url{https://pypi.org/project/ImageHash/}} for the images to compare them. We remove tweets with duplicate images or text, keeping only one instance of each text and image. 
      
    Next, we gather a random sample of tweets collected from the Internet Archive\footnote{\url{https://archive.org}} from around the same time and about the same topics (by using hashtags, this is explained later) as the propaganda tweets to serve as our negative set. They are standardized in the same way as state-sponsored tweets. We remove all potential propaganda tweets from this set by comparing the standardized textual content and the hash of images of tweets collected from the Internet Archive with the state-sponsored tweets. Any tweet containing the same textual or visual content as the state-sponsored tweets are removed from the negative set. Though it is possible that some of these tweets contain unrecognized propaganda, considering the large number of tweets we sample from, their effects will be limited.
    
    For each sub-dataset (dataset from particular organizations), the most important fifteen hashtags of the state-sponsored tweets are chosen as \emph{keywords} and used to filter tweets because of the following two reasons. Firstly, this will ensure that both positive and negative tweets include the same hashtags, which means they are most likely about the same topics. Secondly, it is probably the case that a sizeable number of tweets from the propagandists are innocuous (i.e., not pushing a particular propaganda narrative). These tweets will most likely be about unimportant topics and be used by the propagandists to masqueraded as "regular" users. Thus by focusing on the top hashtags, we are more likely to be collecting actual propaganda tweets as these are the topics that are the main focus of the propagandists.
    
    The importance is calculated as follows:
    \begin{equation}
        C = \sum_{i=1}^N \frac{h_i}{H_i}
    \end{equation}
    where N is the number of state-sponsored tweets for the organization, $h_i$ is the occurrence number of the selected hashtags \textit{h} in the i-\textit{th} tweet, and $H_i$ represents the total number of hashtags of the i-\textit{th} tweet. Only tweets including at least one of those keywords are kept (for both the positive and negative sets). We also randomly sample our tweets so that we have an equal number of tweets in the positive (state-sponsored) and negatives sets for each organization for each month. Therefore, we get a balanced dataset where for each sub-dataset, both state-sponsored and non-state-sponsored tweets are about similar topics. The number of tweets for the final dataset for each organization and their time period is shown in Table \ref{tab:dataset}.

\linespread{1}

    \begin{table}[htb]\scriptsize
        \caption{Dataset details including organizations, number of tweets and time period.}
        \label{tab:dataset}
        \centering
            \begin{tabular}{|l|l|r|l|l|}
            \hline
            Organization             & Data Type       & \multicolumn{1}{l|}{\# of tweets} & Start date & End date   \\ \hline
            \multirow{4}{*}{IRA}     & training        & 4,102                              & 2015-04-01 & 2016-01-19 \\ 
                                     & validation      & 642                               & 2016-01-20 & 2016-01-31 \\ 
                                     & continuous test & 896                               & 2016-02-01 & 2016-02-28 \\ 
                                     & delay test      & 216                               & 2016-10-01 & 2017-06-30 \\ \hline
            \multirow{4}{*}{Russian} & training        & 5,194                              & 2015-04-01 & 2015-12-14 \\ 
                                     & validation      & 742                               & 2015-12-15 & 2015-12-31 \\ 
                                     & continuous test & 1,534                              & 2016-01-01 & 2016-02-28 \\ 
                                     & delay test      & 6,544                              & 2016-10-01 & 2017-06-30 \\ \hline
            \multirow{4}{*}{Iranian} & training        & 5,452                              & 2015-04-01 & 2015-12-14 \\ 
                                     & validation      & 300                               & 2015-12-15 & 2015-12-31 \\ 
                                     & continuous test & 1,620                              & 2016-01-01 & 2016-02-28 \\ 
                                     & delay test      & 6,642                              & 2016-10-01 & 2017-06-30 \\ \hline
            \end{tabular}
    \end{table}
\linespread{0.8}

    Building the dataset from three different organizations allows us to test the generalizability of our framework since different organizations presumably push different narratives in their propaganda and use different strategies in designing their messages. 
    
    Each sub-dataset includes data of two time periods with a gap of seven months. This is because we want to test the generalizability of trained models (i.e. whether a model trained on one time period is predictive on other time periods). By comparing the 15 most common hashtags of these two time periods for the same organization, we find the topics of the tweets during these two time periods are qualitatively different for each organization. Therefore, the delay test data is suitable for testing the temporal generalizability of our model.

    \linespread{1}

    \begin{table}[htb]\scriptsize
    \caption{Example of four different modifications of textual content.}
    \label{tab:text_example}
    \centering
            \begin{tabular}{|l|l|}
            \hline
            Text Type & Textual Content                                                              \\ \hline
            Original  & \#Putin's 1st New Year's "achievement" in \#Syria URL \\ \hline
            Tag       & TAG 1st New Year's "achievement" in TAG URL           \\ \hline
            Miss      & 1st New Year's "achievement" in URL                   \\ \hline
            Structure & T W W W W W T U                                       \\ \hline
            \end{tabular}
    \end{table}
    \linespread{0.8}

     In order to explore the possible use of templates or guidelines used by state-backed propagandists we generate four types of textual content from a tweet: (1) \emph{Original}: the standardized textual content (2) \emph{Tag}: to further study the effect of hashtags on our model, hashtags are replaced with the token 'TAG'. We want to make sure that our model is not just identifying certain combination of hashtags as propaganda (3) \emph{Miss}: to discover whether the number and the position of hashtags in a tweet are predictive for identifying propaganda, hashtags are removed completely (4) \emph{Structure}: to discover whether state-sponsored propaganda follow some kind of structural template, we replace hashtags with the token 'T', URLS with the token 'U', and all other words with the token 'W', to capture the structural template of the tweets. Table \ref{tab:text_example} shows an example tweet and its corresponding modifications.
        
\section{Model}
     This paper aims to comprehensively utilize visual and textual content to determine whether a tweet is state-sponsored propaganda or not. A multi-model neural network is proposed to capture intrinsic relations between textual and visual features of each tweet. This section first provides an overview of our multi-model framework and then explains different sub-networks and the learning method in detail.
    \subsection{Model overview}
\begin{figure}[htb]
        \centering
        \includegraphics[width=1.0\columnwidth]{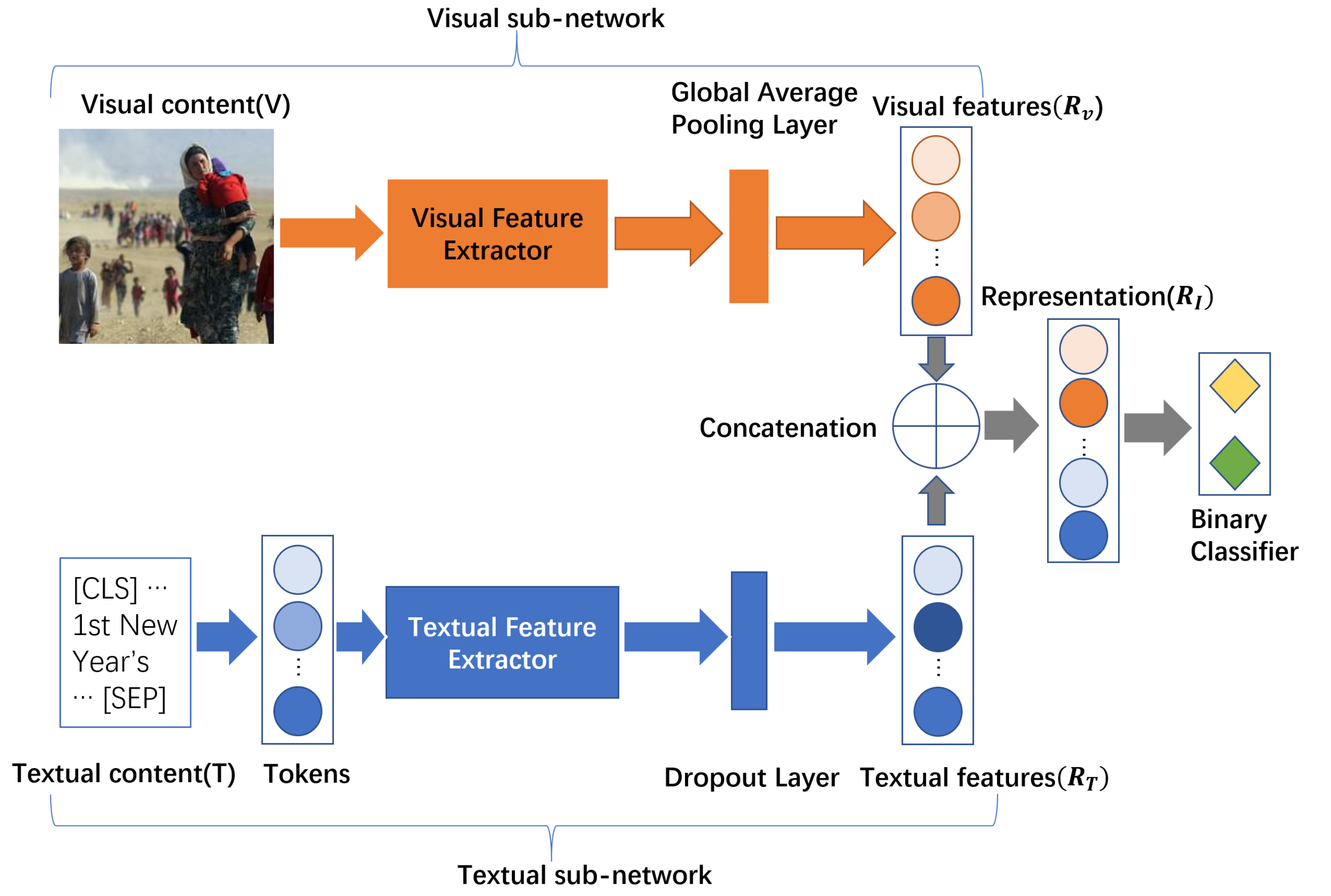}
        \caption{The structure of our multi-model network}
        \label{fig:model}
    \end{figure}

        A tweet instance $I=\{V,T\}$ is defined as a tuple representing two different modalities of content: the visual content \textit{V} and textual content \textit{T}. Our framework extracts features from these modalities and aims to learn a reliable representation $R_I$ as the aggregation of \textit{V} and \textit{T}. The overall structure of our multi-model network is illustrated in Figure \ref{fig:model}. Our framework includes two sub-networks: 1) a visual sub-network that generates visual representation $R_V$ from visual content $V$ and 2) a textual sub-network which learns textual representation $R_T$ from textual content $T$. The visual representation $R_V$ and textual representation $R_T$ are then concatenated to obtain the representation $R_I=[R_V;R_T]$ of the tweet \textit{I}. This representation is then classified by a binary classifier as state-sponsored propaganda or not.
        
    \subsection{Representation of Visual Content}
        The visual sub-network takes an image as input and feeds it to the visual feature extractor which is based on the Convolutional Neural Network (CNN) model to extract features. A global average pooling layer is then used to normalize the extracted features. Because different CNN models focus on features of different aspects, we test seven CNN models: VGG-19, ResNet-50, Inception-v3, Style model, Content model, Style-Content model, and Image-Structure model.
        
        \texttt{ResNet-50} \cite{resnet50}, \texttt{VGG-19} \cite{vgg19} and \texttt{Inception-v3} \cite{inceptionv3} are all designed for the task of image recognition and achieve promising performance. Later application of these models on other image tasks proves that they can also extract predictive features for tasks apart from image recognition. While removing classifier from these networks results in a competitive feature extractor, since these models are all designed for image recognition, other models focusing on different aspects of visual content are also tested for extraction.
        
        Style model, Content model, and Style-Content model are all based on VGG-19. Previous works \cite{gatys2015neural} prove that the representations of style and content in CNN models are partly separable. \emph{Style model} follows the work of image style classification \cite{chu2016deep} to extract feature maps from the thirteenth convolutional layer and calculate gram matrix. \emph{Content model} extracts feature maps from the fourteenth convolutional layer as mentioned in the work of neural style transfer \cite{gatys2015neural}. To extract local features, a convolutional layer activated by relu function is used. To obtain global features, two fully connected layers activated by relu function with 512 and 256 units separately are used to extract features of visual content following the convolutional layer. \emph{Style-Content model} concatenates features extracted by Style model and Content model as new representations of visual content for further prediction. 
        
        \emph{Image-Structure model} focuses on the relationship between text and the objects in the image. As shown in Figure \ref{fig:image_example}, the visual content of tweets may be objects-only, text-only, and the combination of image and objects. As the relationship between objects and text may be predictive for this task, a pre-trained text detector \cite{zhou2017east} is used to derive the pixel level text possibility map. The map is regarded as a grayscale image and fed to an Inception-V3 model to extract structural features.
        
    \subsection{Representation of Textual Content}
        The textual sub-network is designed to extract features from the textual content of tweets. Textual content, after WordPiece tokenization, is fed to a feature extractor to obtain the representation. A dropout layer is used to avoid over fitting.
    
        In order to obtain features from textual content, Bidirectional Encoder Representations from Transformers (BERT) \cite{devlin2018bert} is used as the feature extractor. BERT is based on the transformer architecture \cite{vaswani2017attention} following an encoder-decoder design. While multiple models are tested in the extraction of visual features, the vanilla BERT model is the only one used for extracting textual features because BERT-based models are the most competitive in the textual feature extraction task.
        
        For the given text \textit{T}, following the instruction for fine-tuning BERT \cite{devlin2018bert}, '[CLS]' and '[SEP]' are added before and after of the text, respectively. After being tokenized, the given text is padded with '[PAD]' to obtain a sequence of tokens with fixed length:$T = [t_1,t_2, \cdot,t_n]$, where $t_i$is the i-\textit{th} token in the sentence, and \textit{n} is the sequence length. The sequence is then fed into a $BERT_{LARGE}$ model, which has 24 layers and 16 attention heads for each layer to derive its representation $R_T$. When training, a dropout layer is applied to avoid over fitting.
    \subsection{Model Learning}
        At this point, the representation for visual content $R_V$ and the representation for textual content $R_T$ have been obtained. These two features are concatenated to form the multi-model representation $R_I = [R_V;R_T]$ for the given tweet, which is fed into a binary classifier activated by softmax function before computing its loss. Cross-entropy is employed to define the loss of m-\textit{th} tweets as follows:
        
        \begin{equation}
            p (R_I^m) = softmax (W_cR_I^m+b_c)
        \end{equation}
        \begin{equation}
            L (R_I^m) = -[l^mlogp (R_I^m)+ (1-l^m)logp (1-R_I^m)]
        \end{equation}
        where $R_I^m$ is the multi-model representation of the m-\textit{th} tweet instance, $W_c$, $b_c$ are learnable parameters, and $l^m$ denotes the ground truth label for the m-\textit{th} instance, with 1 representing state-sponsored tweets and 0 representing non-state-sponsored ones. 
        
        In summary, our multi-model framework takes inputting training data $I =\{T; V\}$ with contents from two different modalities: text and image. It outputs the prediction label for each instance to indicate whether the tweet is state-sponsored or non-state-sponsored. The whole model is trained end-to-end with batched Stochastic Gradient Descent to minimize the loss function:
        \begin{equation}
            L = -\frac{1}{N} \sum_{m=1}^N [l^mlogp (R_I^m)+ (1-l^m)logp (1-R_I^m)]
        \end{equation}
        where N is the total number of instances.
        
\section{Experiments and Results}
        All our models are trained on sub-datasets separately and tested on the corresponding continuous and delay test data. In order to avoid over fitting, all models except the text-only model are trained with early-stop policy, and models with the best performance on validation data are tested. Image feature extractors are initialized with the pre-trained weights learnt from ImageNet dataset, and, unless further noted, BERT models are initialized with the pre-trained weights published by Google. Considering that the performance of the same model on different sub-dataset may change a lot, arithmetic mean is used to compare the performance of different models.
    \subsection{Image-only models}
        To justify whether all feature extractors can extract predictive features from visual content, for each model used as visual feature extractor, an image-only network is built by adding a classifier to the visual sub-network. Arithmetic mean performance of three sub-datasets is used to test the image-only model. As shown in Table \ref{tab:image-only-performance}, all models can extract predictive features on continuous test data (accuracy=0.658), but only certain models extract features correctly from delay test data. On both continuous test data and delay test data, ResNet-50 is the best choice.
        \linespread{1}

        \begin{table}[!htb]\scriptsize
        \caption{F1 score of image-only models}
        \label{tab:image-only-performance}
        \centering
            \begin{tabular}{|l|ll|}
            \hline
                                                     & Continuous test data           & Delay test data        \\ \hline
            Resnet-50                                &\textbf{0.714} &\textbf{0.644}  \\ \hline
            Inception-v3                             &0.706 &0.639  \\ \hline
            VGG-19                                   &0.681 &0.497  \\ \hline
            Style                                    &0.477 &0.424  \\ \hline
            Content                                  &0.684 &0.519  \\ \hline
            Texture-Content                          &0.664 &0.510  \\ \hline
            Image-Structure                          &0.604 &0.446  \\ \hline
            \end{tabular}
        \end{table}
        \linespread{0.8}
    \subsection{Text-only models}
        To justify whether hashtags and the structure of textual content influence the identification of state-sponsored tweets, text-only models are built by adding a classifier to the textual sub-network. As mentioned in the Data section, \emph{Original} type, \emph{Tag} type, \emph{Miss} type and the \emph{Structure} type of textual content are used to train and test text-only models. As shown in Table \ref{tab:text-only-performance}, all types of textual content provide predictive features for identifying state-sponsored tweets on continuous test data (accuracy=0.772) and most types of textual content provide meaningful features on delay test data. For continuous test data, Original type is the best choice of textual content (accuracy=0.861 and f1=0.854). For delay test data, Tag type is the best choice for textual content (accuracy=0.641 and f1=0.649).
        \linespread{1}
        
        \begin{table}[!htb]\scriptsize
        \caption{F1 score of text-only model. Continuous means continuous test dataset and Delay means delay test dataset}
        \label{tab:text-only-performance}
        \centering
        \begin{tabular}{|l|l|l|}
        \hline
                  & Continus test data & Delay test data \\ \hline
        Original  & \textbf{0.854}     & 0.643           \\ \hline
        Tag       & 0.803              & \textbf{0.649}  \\ \hline
        Miss      & 0.788              & 0.614           \\ \hline
        Structure & 0.715              & 0.548           \\ \hline
        \end{tabular}
        \end{table}
        \linespread{0.8}
    \subsection{Multi-model models}
         All combinations of different image feature extractors and different types of textual content are tested to discover the best choice for the task. As shown in Table \ref{tab:fused-continuous-performance} and Table \ref{tab:fused-delay-performance}, all combinations achieve significant performance on continuous test data, and most combinations achieve promising performance on delay test data. 
        
        For continuous test data, as shown in Table \ref{tab:fused-continuous-performance}, the combination of Original type of textual content and ResNet-50 is the best choice (accuracy=0.879 and f1=0.869). For delay test data, as shown in Table \ref{tab:fused-delay-performance}, the combination of Miss type of textual content and ResNet-50 achieves the best performance (accuracy=0.696 and f1=0.697). 
        \linespread{1}

        \begin{table}[!htb]\scriptsize
        \caption{F1 score of models on continuous test data.}
        \label{tab:fused-continuous-performance}
        \centering
            \begin{tabular}{|l|llll|l|}
            \hline
                                                          & Original           & Tag        & Miss    & Structure   &Average\\ \hline
            BERT+Resnet-50                                &\textbf{0.869}      &0.836       &0.813    &0.796        &0.828  \\ \hline
            BERT+Inception-v3                             &0.865               &0.838       &0.806    &0.770        &0.820  \\ \hline
            BERT+VGG-19                                   &0.862               &0.811       &0.802    &0.766        &0.810  \\ \hline
            BERT+Style                                    &0.864	           &0.797	    &0.789	  &0.736	    &0.797  \\ \hline
            BERT+Content                                  &0.867	           &0.815	    &0.801	  &0.742	    &0.806  \\ \hline
            BERT+Texture-Content                          &0.861	           &0.817	    &0.805	  &0.759	    &0.811  \\ \hline
            BERT+Image-Structure                          &0.861	           &0.794	    &0.780	  &0.742	    &0.794  \\ \hline
            \end{tabular}
        \end{table}
        \begin{table}[!htb]\scriptsize
        \caption{F1 score of models on delay test data.}
        \label{tab:fused-delay-performance}
        \centering
            \begin{tabular}{|l|llll|l|}
            \hline
                                                          & Original   & Tag        & Miss    & Structure   &Average  \\ \hline
            BERT+Resnet-50                                &0.684	   &0.695	    &\textbf{0.697}	  &0.684	    &0.690     \\ \hline
            BERT+Inception-v3                             &0.618	   &0.679	    &0.647	  &0.632	    &0.644\\ \hline
            BERT+VGG-19                                   &0.619	   &0.639	    &0.599	  &0.555	    &0.603   \\ \hline
            BERT+Style                                    &0.609	   &0.639	    &0.613	  &0.549	    &0.602   \\ \hline
            BERT+Content                                  &0.605	   &0.642	    &0.604	  &0.555	    &0.602  \\ \hline
            BERT+Texture-Content                          &0.627	   &0.618	    &0.609	  &0.541	    &0.599  \\ \hline
            BERT+Image-Structure                          &0.539	   &0.612	    &0.636	  &0.532	    &0.580  \\ \hline
            \end{tabular}
        \end{table}
        \linespread{1}

    \subsection{Generalizability across sub-dataset}
        
        To discover whether the generalizability of models can be improved by inhibiting features relevant to organizations, we train our models on the data of two organizations and test on the other organization (e.g. train on IRA and Iranian-backed data and test on Russian-linked data). For each organization, we test the performance of models trained on the other two organizations and use the arithmetic mean of their performance as the baseline (e.g. We have two models one is trained on IRA data and the other is trained on Iranian-backed data. We test them on Russian-linked data and use the arithmetic mean of them as the baseline of Russian-linked data). For both baseline and models trained on multi organizations, we train the combination of Resnet-50 and Original textual content and the combination of Resnet-50 and Miss textual content because their best performance on continuous and delay test data. As shown in Table \ref{performance-generalizability}, compared with the baseline, models trained on tweets from multi organizations achieve better performance. Especially the models trained on IRA and Iranian sub-datasets, which achieves the accuracy of 0.615 on delay test data, is very close to the best performance of models trained on Russian sub-dataset whose accuracy is 0.643.
        \linespread{1}

        \begin{table}[!htb]\scriptsize
        \caption{Generalizability of models with F1-score. Baseline refers to the baseline models, and Multi means the model trained on two sub-dataset. (C) refers to the continuous test data, and (D) refers to the delay test data.}
        \label{performance-generalizability}
        \centering
        \begin{tabular}{|llll|}
        \hline
                            & IRA    & Russian-linked & Iranian-backed \\ \hline
        baseline (C) & 0.382 & 0.170         & 0.453          \\ 
        multi (C)    & \textbf{0.567} & \textbf{0.295}         & \textbf{0.629}         \\ \hline
        baseline (D)      & 0.356 & 0.2991        & 0.400            \\
        multi (D)         & \textbf{0.524} & \textbf{0.535}         & \textbf{0.613}         \\
        \hline
        \end{tabular}

        \end{table}
        \linespread{0.8}

\section{Discussion}
    To reduce the potential damage of type \uppercase\expandafter{\romannumeral2} error, we: 1) conduct error analysis especially change the threshold to analysis the relationship between false positive rate and true positive rate; 2) use visualization tools to shed light on the mechanism of our framework to discover the potential explanations for the decisions made by our models. To study the generalizability of features and the relationship between generalizability and the performance of models, we analysis the models trained on different input data and discover potential factors of generalizability and performance.

    \subsection{Error analysis}
        
        \begin{figure}[!htb]
        \centering
            \subfloat[Continuous test data of IRA]{
            \includegraphics[width=0.47\columnwidth]{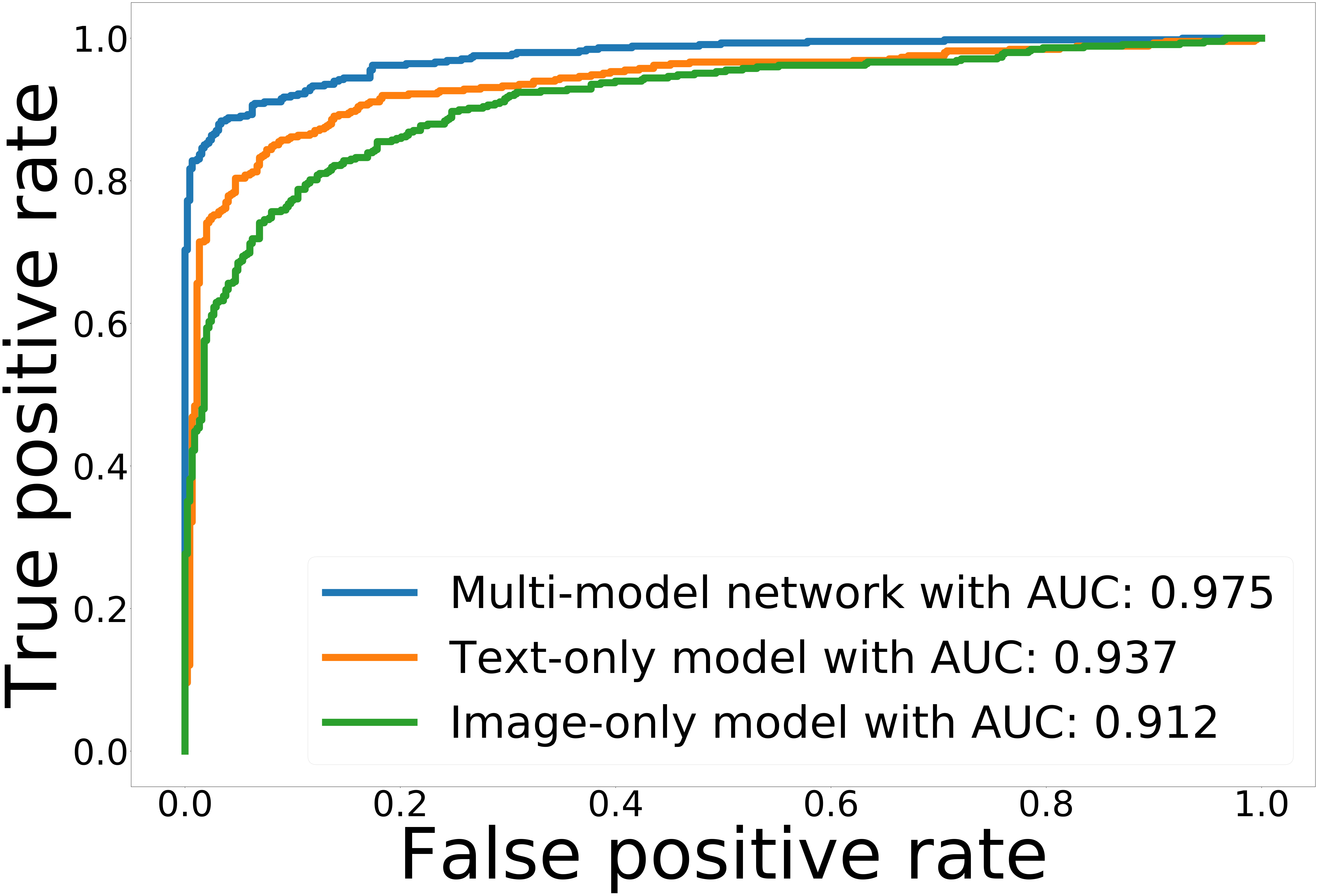}
            \label{fig:c_ira}
            }
            \subfloat[Delay test data of IRA]{
            \includegraphics[width=0.47\columnwidth]{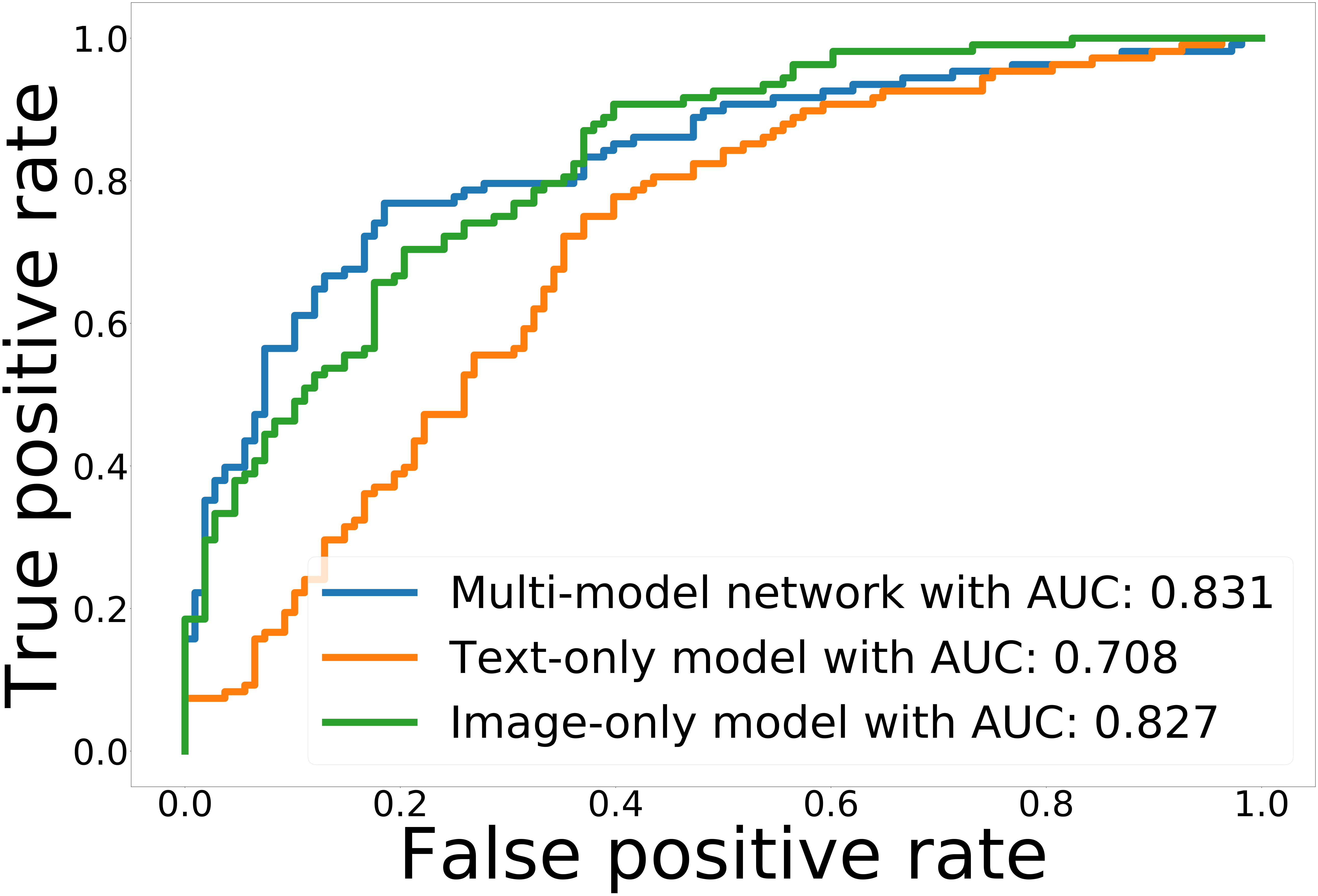}
            \label{fig:d_ira}
            }
            
            \subfloat[Continuous test data of Russian]{
            \includegraphics[width=0.47\columnwidth]{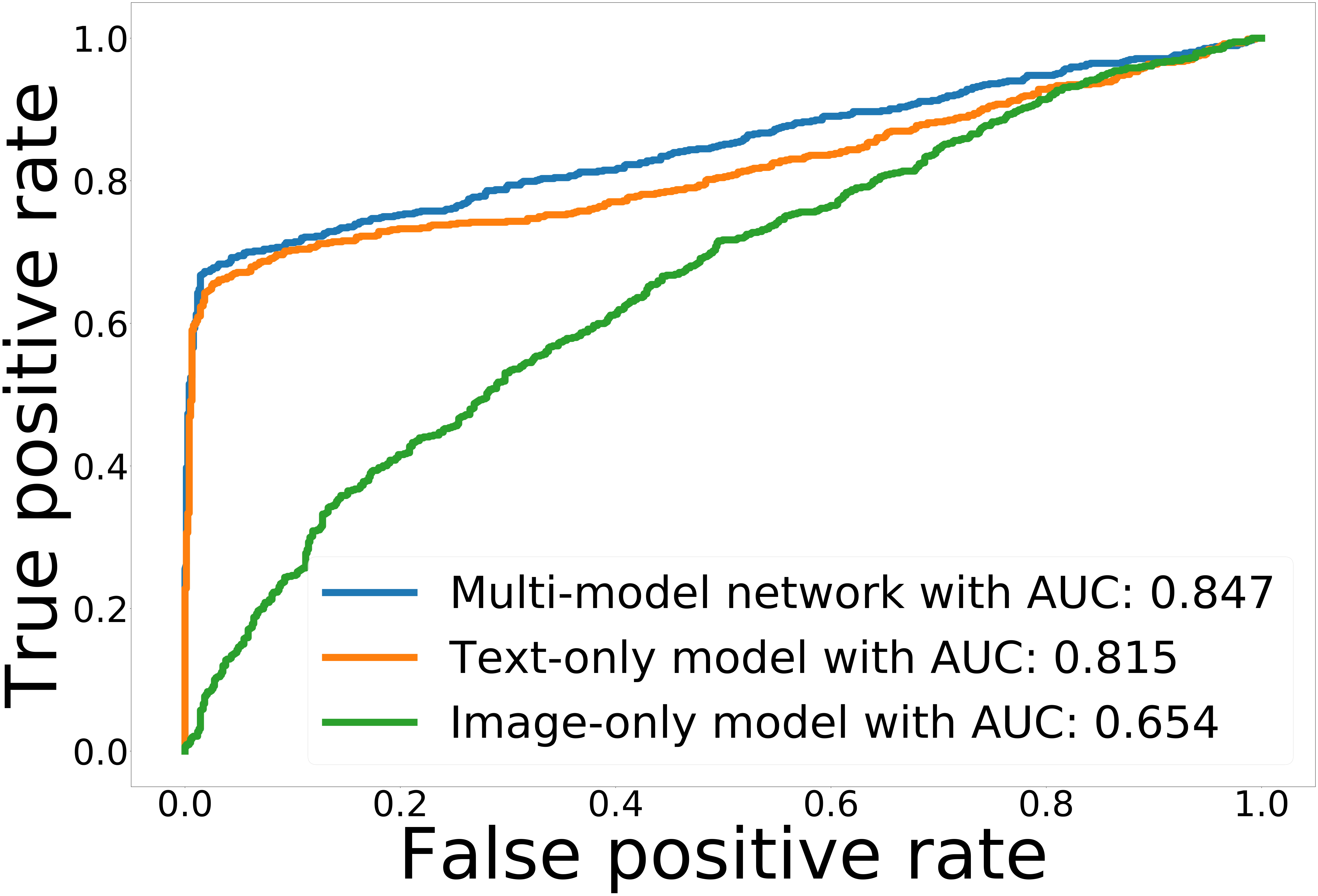}
            \label{fig:c_russian}
            }
            \subfloat[Delay test data of Russian]{
            \includegraphics[width=0.47\columnwidth]{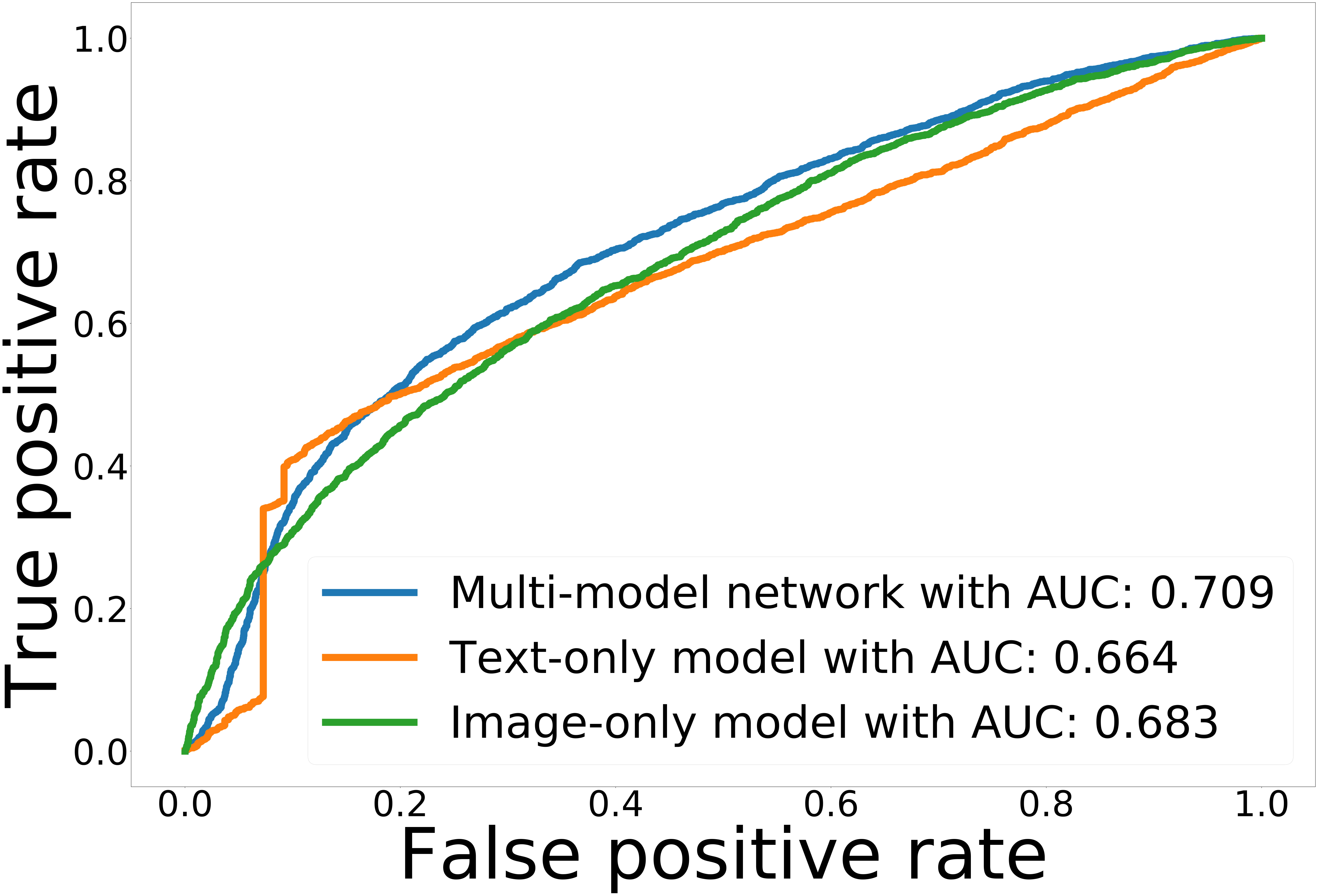}
            \label{fig:d_russian}
            }
            
            \subfloat[Continuous test data of Iranian]{
            \includegraphics[width=0.47\columnwidth]{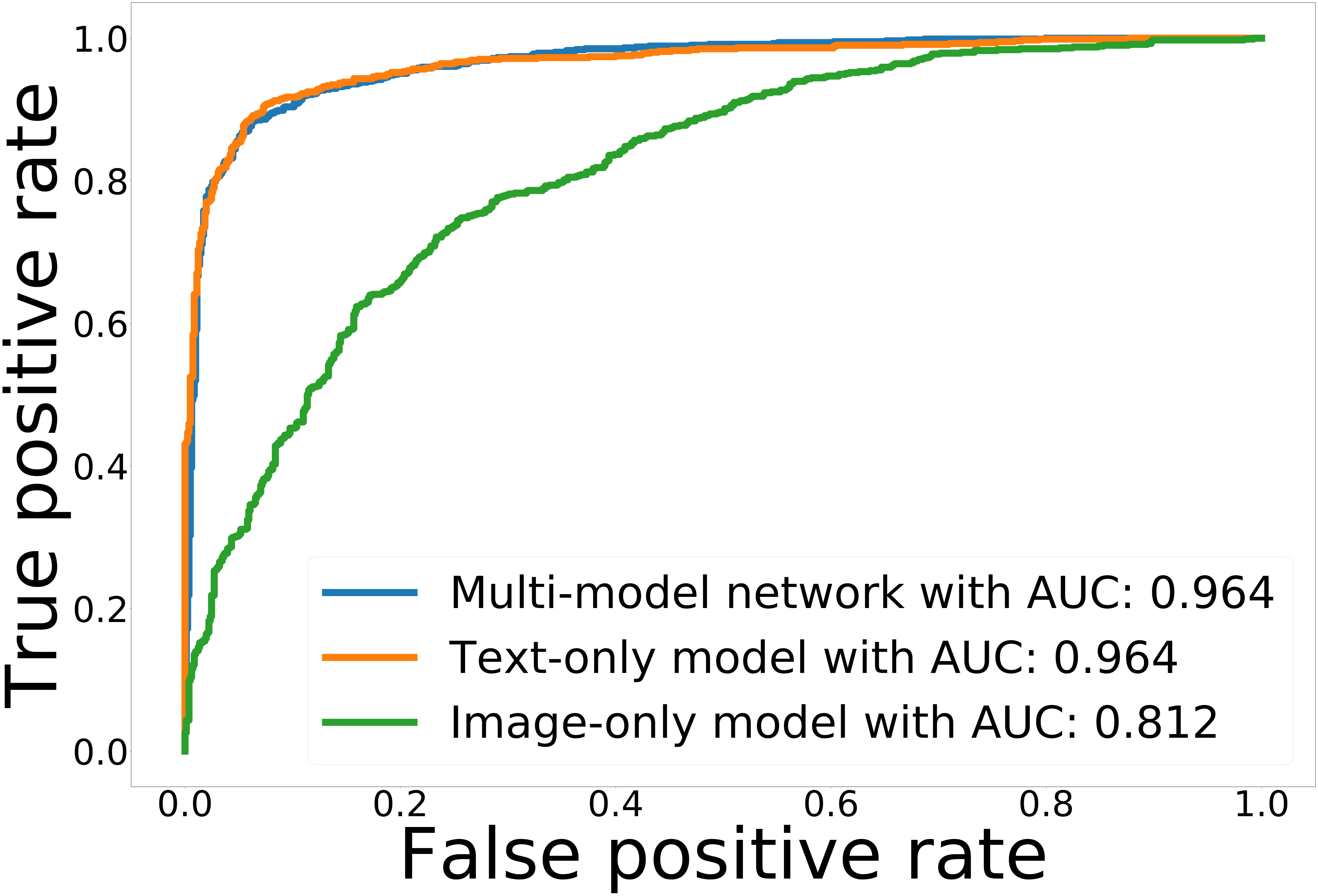}
            \label{fig:c_iranian}
            }
            \subfloat[Delay test data of Iranian]{
            \includegraphics[width=0.47\columnwidth]{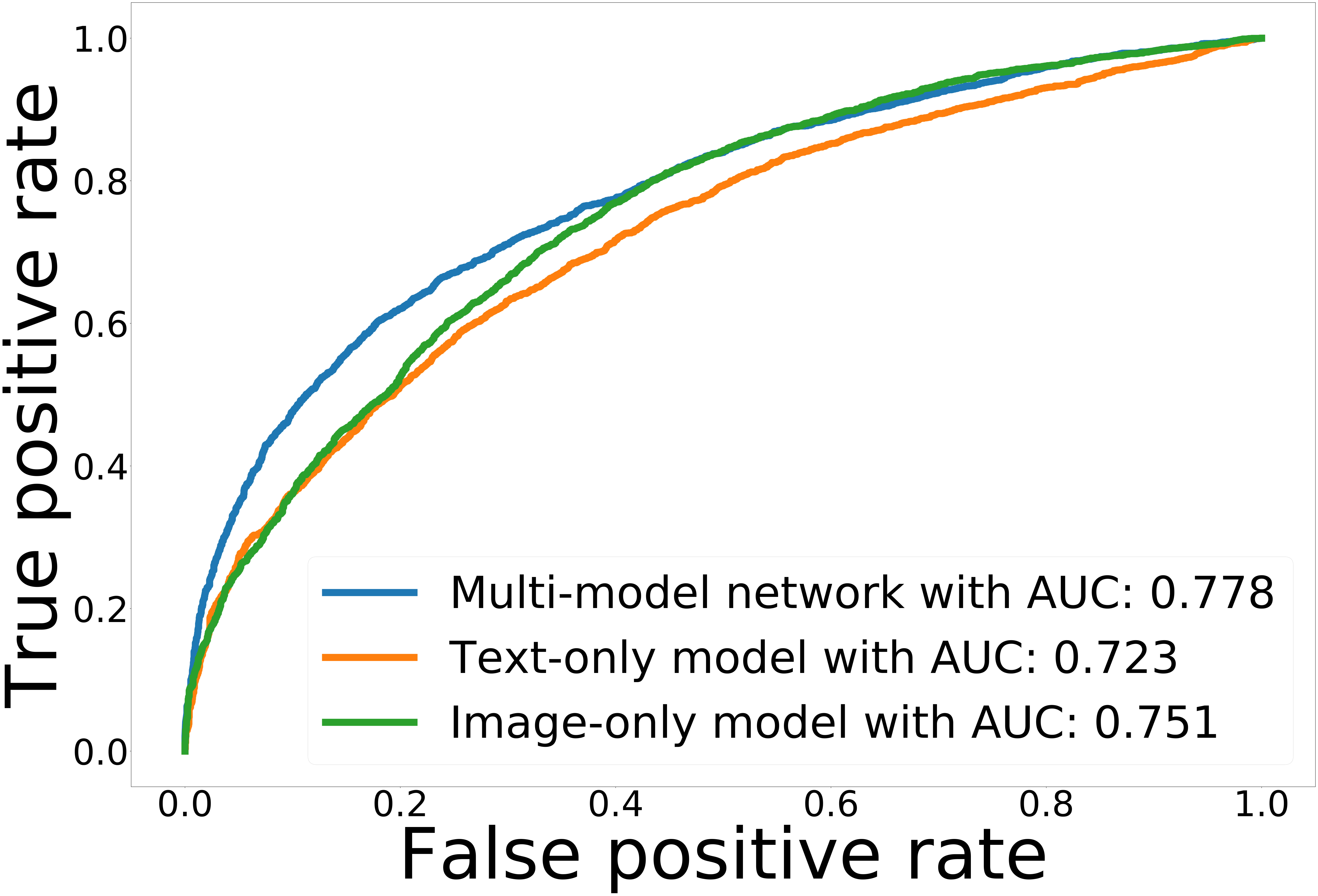}
            \label{fig:d_iranian}
            }

        \caption{ROC curve and AUC for picked models on both Continuous and Delay test dataset}
        \label{fig:roc_curve}
        \end{figure}
        
        Receiver Operating Characteristic curve (ROC curve) and Area Under Curve (AUC) are used as metrics for error analysis. We compare the ROC curve and AUC of the multi-model network which achieves best performance on continuous and delay test data (i.e. the combination of ResNet-50 and Original textual content on continuous test data and the combination of ResNet-50 and Miss textual content on delay test data) and its corresponding image-only and text-only models.
        
        As shown in Figure \ref{fig:roc_curve}, the performance of the multi-model network (blue) is better than that of the image-only (green) and the text-only model (orange). The performance on continuous is much better than that on delay test data proves that the change of topics influences our models significantly.

        \linespread{1}

        \begin{table}[h]\scriptsize
            \caption{True positive rate when false positive rate is 0 for picked models}
            \centering
            \begin{tabular}{|l|l|lll|}
            \hline
            Organization             & Dataset    & Multi-model & Image-only & Text-only \\ \hline
            \multirow{2}{*}{IRA}     & continuous & \textbf{0.703}       & 0.696      & 0.277     \\ 
                                     & delay      & 0.157       & 0.074      & \textbf{0.185}     \\ \hline
            \multirow{2}{*}{Russian} & continuous & \textbf{0.256}       & 0.228      & 0.007     \\ 
                                     & delay      & 0.000       & 0.000      & 0.000     \\ \hline
            \multirow{2}{*}{Iranian} & continuous & 0.041       & \textbf{0.431}      & 0.026     \\ 
                                     & delay      & 0.000       & 0.000      & 0.010     \\ \hline
            \end{tabular}
            \label{tab:threshold}
        \end{table}
        \linespread{0.8}

        Compared with false negative instances (i.e. labelling state-sponsored tweets as non-state-sponsored), we are concerned more about false positive instances because, when models are applied in the real world, the false positive instances are likely to limit the expression of users and their right of free speech. Therefore, we change the threshold to compare the true positive rate of different models when the false positive rate is 0. As shown in Table \ref{tab:threshold}, the multi-model network and the image-only network achieve promising true positive rate on continuous test data. Of particular note is that the true positive rate of the multi-model network on IRA data is 0.703, and the average true positive of Image-only model across all three datasets is 0.452. However, no model achieves an acceptable true positive rate on delay test data except for the IRA dataset with 0.157 for the multi-model network and 0.185 for the text-only model. The phenomena prove that the state-sponsored tweets exhibit some exclusive features compared with non-state-sponsored tweets on similar topics. 

    \subsection{Visualization analysis}
        Visualization tools, which highlight the predictive information of input data, provide us an opportunity to understand deep learning models and discover the situations with high incorrectly prediction probability. The examples are generated by the combination of ResNet-50 and Original textual content on continuous test data.
    
        For the visual content, Gradient-weighted Class Activation Mapping (Grad-Cam) \cite{selvaraju2017grad} is a useful technique, producing ‘visual explanations’ for decisions from CNN-based models to render the models more transparent and explainable. Grad-Cam uses the gradients of any target concept flowing into the final convolutional layer to produce a coarse localization map highlighting the important regions in image
       for prediction. 

\begin{figure}[!htb]
            \centering
            \subfloat[True positive example]{
            \includegraphics[width=0.45\columnwidth]{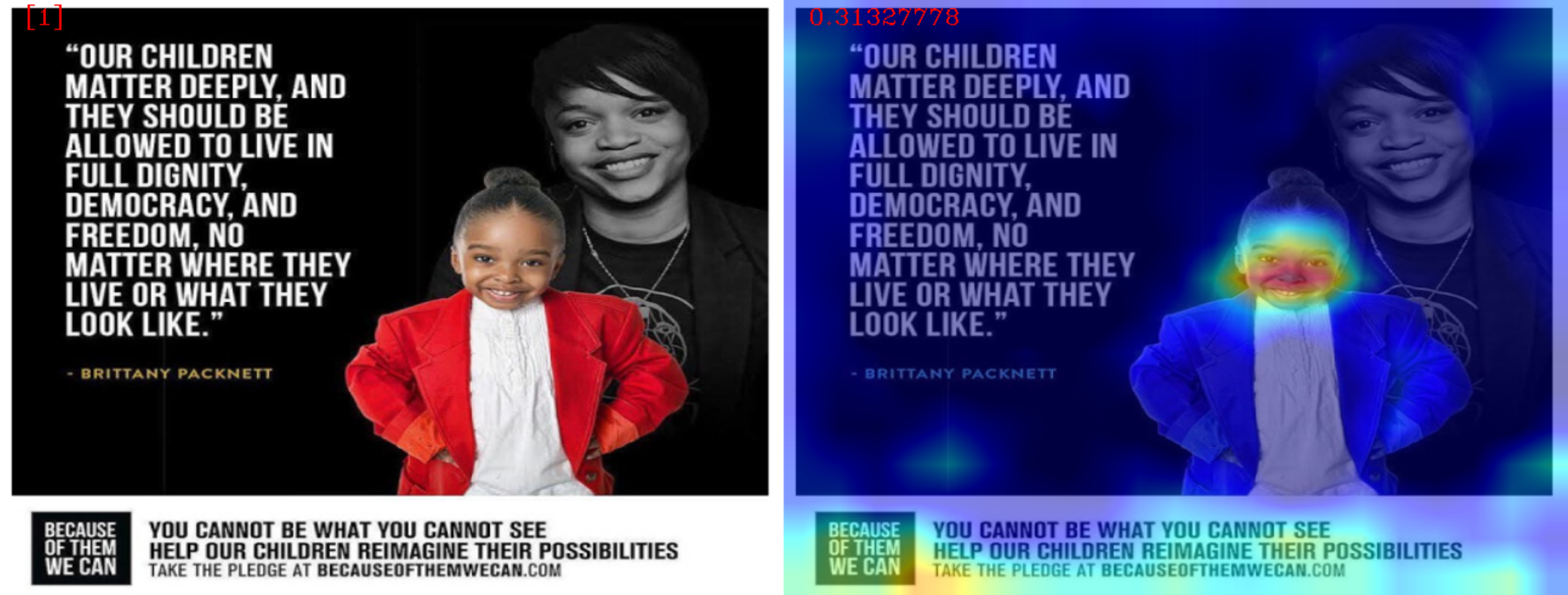}
            \label{fig:TP_image}
            }
            \subfloat[True negative example]{
            \includegraphics[width=0.45\columnwidth]{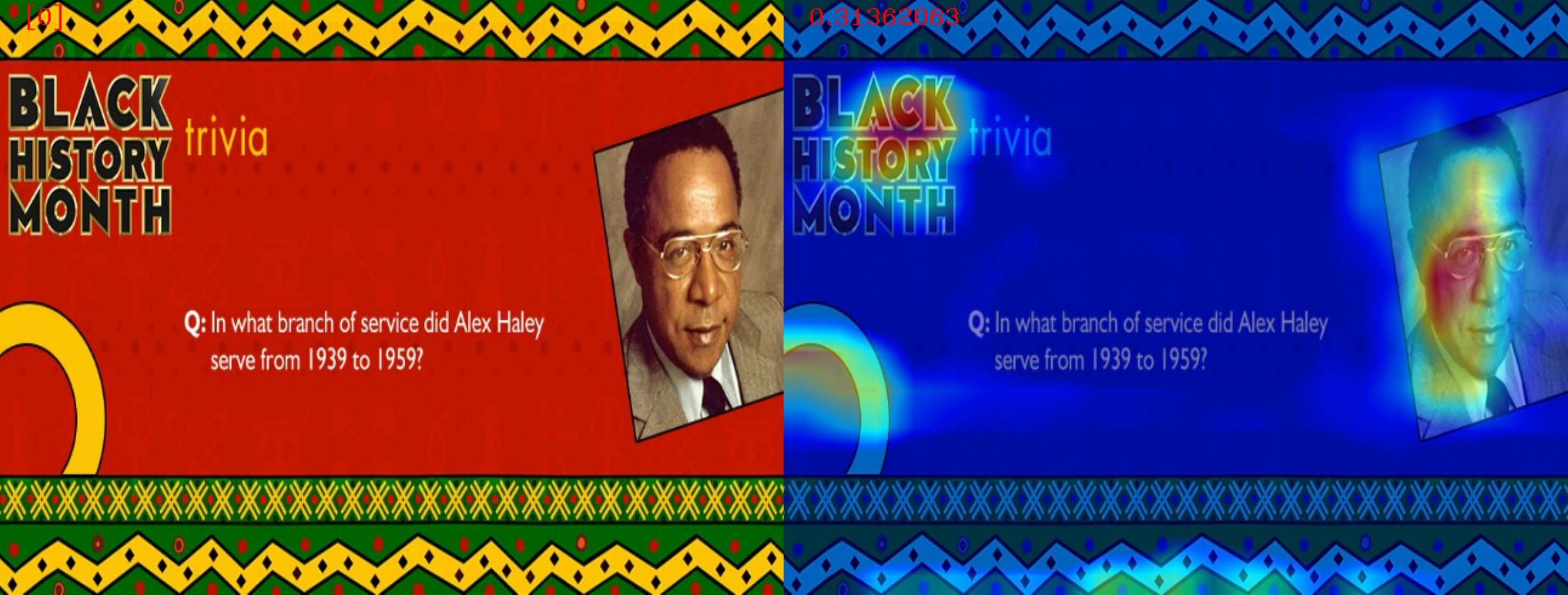}
            \label{fig:TN_image}
            }
            
            \subfloat[False positive example]{
            \includegraphics[width=0.45\columnwidth]{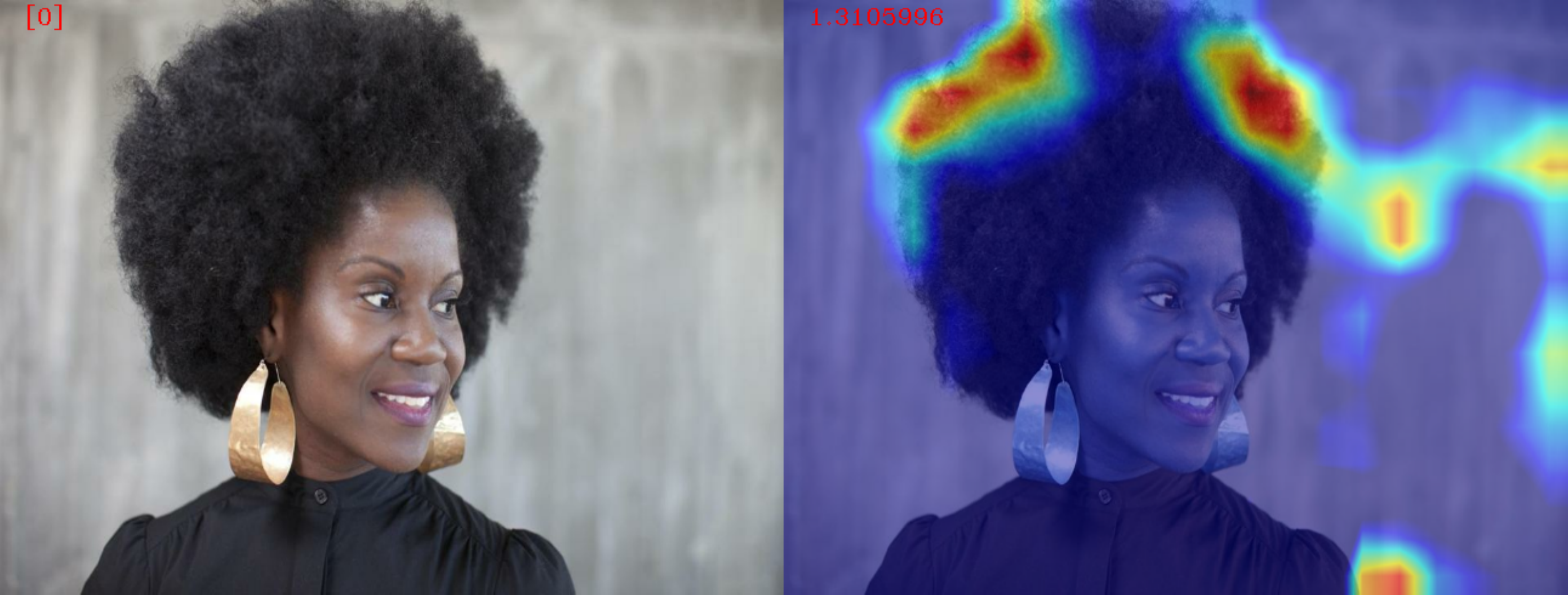}
            \label{fig:FP_image}
            }
            \subfloat[False negative example]{
            \includegraphics[width=0.45\columnwidth]{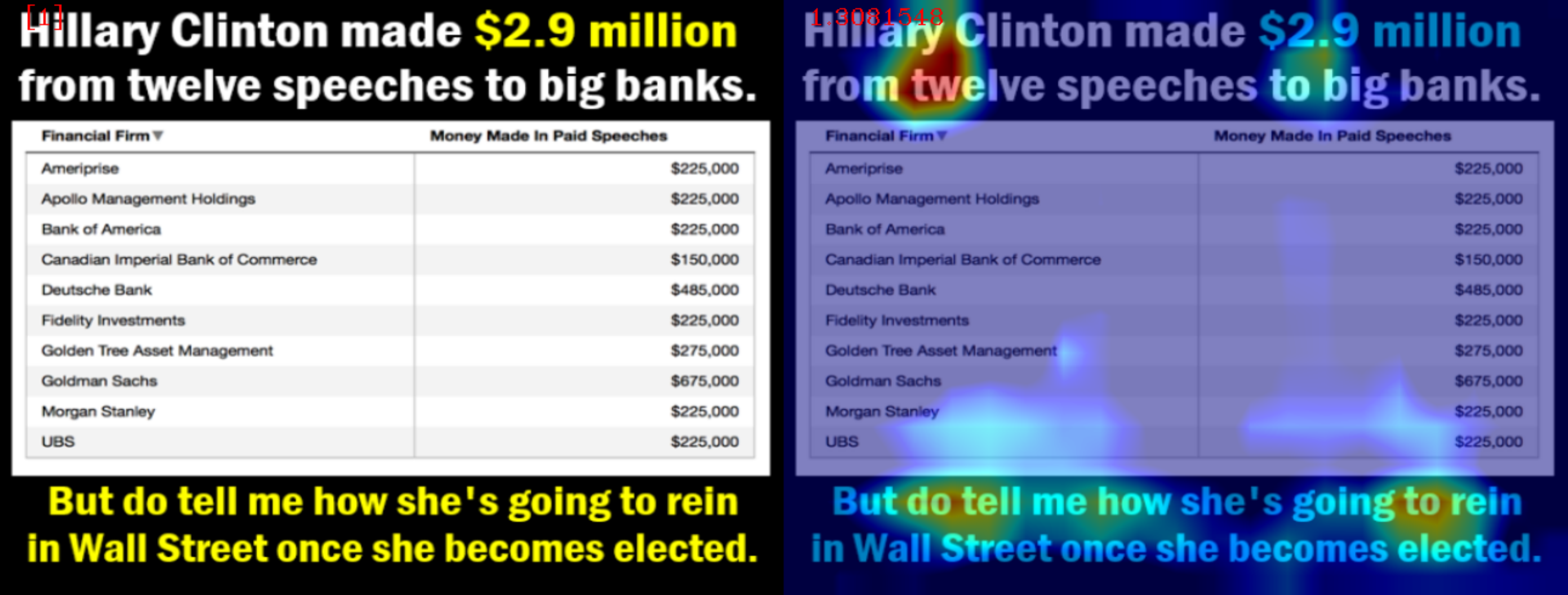}
            \label{fig:FN_image}
            }
            \caption{Examples of Grad-Cam of the visual content. State-sponsored propaganda are labelled as positive. For each example, the left one is the original picture and the right one is the Grad-Cam result}
            \label{fig:image_cases}
        \end{figure}

        By analyzing and comparing the Grad-Cam of correctly and incorrectly predicted examples, we discover the correctness of the prediction is closely related with whether the important region identified by our model is corresponding with social media users' attention. Fig. \ref{fig:image_cases} is used as examples to explain. When given the True positive example (Fig. \ref{fig:TP_image}), user attention is likely to be attracted by the girl and the text at the bottom, which our model also emphasizes. The same phenomenon can be noticed in the True negative example (Fig. \ref{fig:TN_image}). When given the image, users are likely to focus on the text on the top left and the portrait, which are also given high importance by our model. Meanwhile, if the model labels something unattractive as important or ignores certain predictive factors, predictions are likely to be incorrect labelled. For example, in False positive example (Fig. \ref{fig:FP_image}), our model put emphasis on the background while ignoring the face and the earring, and in the False negative example (Fig. \ref{fig:FN_image}), our model highlights insignificant words such as 'tell' and 'twelve' while ignoring important words such as ’Hillary Clinton’ and ’Wall Street’.
        
        For the textual content, the attention map of the last layer of BERT, representing the attention weight from the token being updated to the token being attended to, is visualized. The sentence is de-tokenized to transfer the token-level attention map to more intuitive word-level attention map. The attention from '[SEP]' to other words is set to 0. The importance of each word is calculated by summing up the attention pointing to it and normalized by softmax function. Word '[CLS]' is the mark for the beginning of the sentence. Word '[SEP]' is the mark for the end of the sentence and represents all the attention pointing to the word '[PAD]'.

        From the attention map of the textual content, we notice that hashtags play an important role for both correctly and incorrectly predicted examples. By comparing the importance score, it's noticeable that '[CLS]' which represents the whole sentence is given high importance for positive labels, while certain meaningful words (e.g. '\#BlackHistoryMonth', 'CET/JKT', '\#HillaryCliton') will be given high importance for negative labels. This may partly explain why identifying state-sponsored propaganda is challenging for users because this task depends more on the whole sentence instead of individual words.
        
        In order to exploit the relationship between highlighted regions of visual and textual content, we analyze visual and textual content from the same tweets. 
        
        This analysis demonstrates that our models are likely to predict incorrectly in two situations. The first is when neither of the sub-networks focus on predictive information. For instance, in the false positive example, the visual sub-network focuses on the hair and the background instead of the face and earrings, while the textual sub-network focuses on the '[CLS]'. The second is when the attention of visual content and textual content is not harmonious. For example, in the false negative example, '\#HillaryClinton' is the most important word in textual content, while the visual sub-network ignores it and gives attention to its neighboring words. 
        
        The visualization analysis sheds light on the predictions made by our model, thus allowing us to better interpret the results. This is specially useful when attempting to understand wrong predictions made by our model.
        
    \subsection{Feature generalizability analysis}
    
        The generalizability of a model depends on three types of key features extracted and utilized by the model: 1) fundamental features, which can be used as finger-prints to identify state-sponsored tweets; 2) organization features, which are closely related to the specific propagandist; and 3) topic features, which are related not only with propagandists but also with topics. An appropriate combination of these three types of features is required to reach a point of equilibrium between generalizability and the performance of our models.
        
        In our models, fundamental features are masked by organization and topic features which means that our models cannot identify state-sponsored tweets from other organizations. Organization features are extracted by our model and used for identification, as demonstrated by the fact that most models achieve acceptable performance on delay test data (shown in Table \ref{tab:fused-delay-performance}). Compared with organization features, topic features are prior in our models. This is supported by the better performance of our models on continuous test data (shown in Table \ref{tab:fused-continuous-performance}) compared to delay test data (shown in Table \ref{tab:fused-delay-performance}). This shows that models that put emphasis on topic features will perform better when training and testing data are similar, though limiting their generalizability.
        
        We use text-only models to show the relationship between organization and topic features (see Table \ref{tab:text-only-performance}). Original textual content, which can provide more predictive topic features due to the inclusion of hashtags, leads to better performance of  models on continuous test data compared with other types of textual content. However, when testing on delay test data, models trained on tag content achieve better performance because they can extract more organization features. 
        
        The performance of text-only models based on structure textual content deserves extra attention. Due to a lack of meaningful content, most of the features extracted and utilized by these models should be organization features. As shown in Table \ref{tab:text-only-performance}, the difference between performance on continuous test data and delay test data is the lowest among all text-only models. This phenomenon proves that we can utilize organization features to identify state-sponsored tweets, but the performance of the models will be negatively influenced by the lack of predictive topic features.
        
        Achieving a model that allows for a higher level of generalizability and achieving a higher performance model is a trade-off. As shown in Table \ref{performance-generalizability}, when training with the data of two organizations, our models can identify tweets from other organizations at the cost of reducing the performance on data from the training organization. The phenomenon is because when training with the data of multiple organizations, our models are more likely to learn fundamental features, which can provide the model with the highest level of generalizability, and inhibit the organization features, which can increase the performance of models on topic-relevant tweets from the same organization.

\section{Conclusion}
    In this paper, we present a framework for identifying state-sponsored propaganda on Twitter. The framework uses a multi-model neural network and relies only on textual and visual content, which can be collected immediately after a tweet is published. The multi-model neural network is end-to-end, allowing us to predict whether a tweet is state-sponsored propaganda by directly analyzing the message without the need for extra information (such as user information, social context, bot usages, etc). Our models can accurately separate state-sponsored tweets from non-state-sponsored tweets on similar topics. More importantly, these models function effectively even on data published eight month after the training data. We also attempt to bring transparency to the inner workings of our model by shedding light on the predictive features used in our model to obtain a deeper understanding of the task. 
    
    Overall, our results show that state-sponsored propaganda is not identical to non-state-sponsored posts on Twitter. We find that state-sponsored propaganda have unique characteristics on fundamental, organizational and topic levels. Although these characteristics vary across different propagandists, they can be learnt by a universal framework and leveraged to identify state-sponsored propaganda on social media. Because these characteristics are not hand-crafted, it is difficult for propagandists to manipulate or conceal them.
    
    This paper presents a push in the direction of understanding and identifying state-sponsored propaganda on social media, which is a critical challenge in the race towards curbing the spread of state-sponsored propaganda. In our future endeavors, we aim to validate our framework on other known state-sponsored propaganda organizations. We also aspire to provide a network that can identify different propagandists which requires the extraction of more fundamental characteristics.

\bibliographystyle{IEEEtran}
\bibliography{main.bib}
\end{document}